\pdfoutput=1

\documentclass[11pt]{article}

\usepackage[]{acl}

\usepackage{times}
\usepackage{latexsym}
\usepackage{adjustbox}
\usepackage[T1]{fontenc}

\usepackage[utf8]{inputenc}

\usepackage{microtype}

\usepackage{CJKutf8}
\usepackage{graphicx}
\usepackage{caption}
\usepackage{multirow}
\usepackage{booktabs}
\usepackage{mathtools}
\usepackage{amsmath}
\usepackage{amsfonts}
\usepackage{makecell}
\usepackage{xspace}   
\newcommand{\sep}{\textsc{[sep]}\xspace} 
\newcommand{\pad}{\textsc{[pad]}\xspace} 
\newcommand{\cls}{\textsc{[cls]}\xspace} 

\title{Roof-Transformer: Divided and Joined Understanding with Knowledge Enhancement}

\author{Wei-Lin Liao \\
  Academia Sinica \\
  \\\And
  Cheng-En Su \\
  Academia Sinica \\
  \texttt{\{willsonliao, jimmysu, ma\}@iis.sinica.edu.tw}
  \\ \And
  Wei-Yun Ma \\
  Academia Sinica \\
  }

\begin{document}
\begin{CJK}{UTF8}{gbsn}
\maketitle
\begin{abstract}
Recent work on enhancing BERT-based language representation models with
knowledge graphs (KGs) and knowledge bases (KBs) has yielded promising results on
multiple NLP tasks. State-of-the-art approaches typically integrate the
original input sentences with KG triples and feed the combined
representation into a BERT model. However, as the sequence length of a BERT
model is limited, such a framework supports little knowledge other than the   
original input sentences and is thus forced to discard some knowledge. This
problem is especially severe for downstream tasks for which the input is a long
paragraph or even a document, such as QA or reading comprehension tasks. We
address this problem with Roof-Transformer, a model with two underlying BERTs
and a fusion layer on top. One underlying BERT encodes the knowledge
resources and the other one encodes the original input sentences, and the
fusion layer integrates the two resultant encodings. Experimental results
on a QA task and the GLUE benchmark attest the effectiveness of the proposed model.
\end{abstract}

\section{Introduction}
\label{sec1}
Although BERT dominates multiple benchmark datasets, various studies have conducted to
incorporate extra knowledge into language models (LMs) to advance language 
modeling~\cite{ERNIE, KBERT-2019, KEPLER}. Sources of this
extra knowledge are mostly knowledge graphs (KGs) and knowledge bases (KBs)
that contain rich knowledge facts and benefit language understanding. 
ERNIE~\cite{ERNIE}, for example, employs TransE~\cite{TransE} to  encode entity
information, and concatenates them with token embeddings for input to a
fusion layer. Despite its success on the GLUE benchmark, ERNIE does not 
consider textual knowledge representation due to its token-level concatenation.
K-BERT~\cite{KBERT-2019}, converts knowledge triples into
textual forms and injects them into the input sentences, forming a tree
representation for input to BERT. 
However, this approach considers little knowledge
other than the input sentences due to BERT's intrinsic input length limitation 
(512 tokens). 

Accordingly, we propose Roof-Transformer, a model with two underlying BERTs and 
  a fusion layer, the Transformer encoder~\cite{Transformer} as a fusion layer   
acting as a ``roof'' over the two BERTs. 
Roof-Transformer encodes the text input with one of the
underlying BERTs and uses the other BERT to encode the knowledge information, and
then integrates the two embeddings with the fusion layer for downstream tasks. 
This structure allows us to incorporate information from both the original
text and from external knowledge sources.  
In addition, if memory permits and long input is needed,
more than two BERTs can be employed using this structure.

\begin{figure*}[t!]
  \makebox[\textwidth][c]{\hspace{+6.5em}\includegraphics[width=0.85\textwidth]{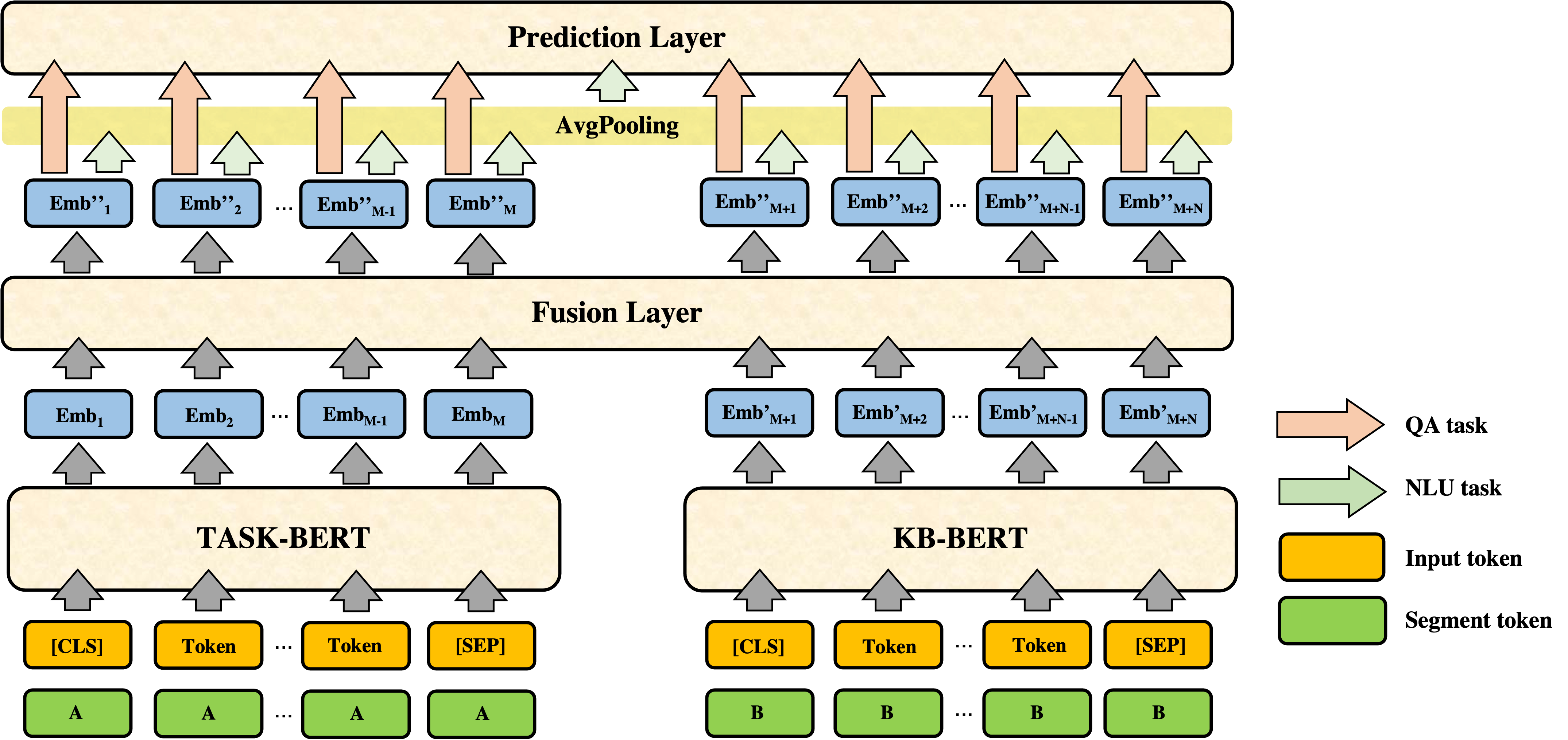}}
  \caption{The overall architecture of Roof-Transformer: The input of Roof-Transformer contains input tokens and segmentation tokens. In the case of QA task, the Prediction Layer directly takes the output of Fusion Layer as input; while in the case of NLU tasks in GLUE, the Prediction Layer takes the mean of the output of Fusion Layer as input.}
  \label{fig:fig1}
\end{figure*}

Despite the intuitive nature of the proposed idea, several critical
challenges must to be addressed:

(1) What is an appropriate model for a ``roof'', and how does the roof distinguish
individual outputs from the underlying BERTs?

(2) How many layers are needed for the roof to fuse the outputs from the
underlying BERTs? This could involve a trade-off between computational resources
and performance.

(3) Various model complexities may result in convergence times for the roof that 
differ from those of the BERTs. How should we address this during training?

(4) The proposed architecture allows for up to 512 tokens of knowledge. 
Although long, this remains a limitation; thus precise knowledge selection
and effective representation are crucial to ensure good performance. 

We investigate various factors and propose solutions for these
challenges, as described in the following sections.

We conduct experiments on the QA task~\cite{QA} with Chinese KBs and the GLUE benchmark with English knowledge corpus. Experimental
results reveal that integrating knowledge using Roof-Transformer 
outperforms using a single BERT to integrate both the original input
sentences and 
external 
knowledge.

Our contributions are summarized as follows:

\vspace{-1mm}
\begin{itemize}
    \itemsep -1pt {}
	 \item We propose Roof-Transformer, an architecture which uses 
	 two distinct BERTs to encode knowledge and input sentences,
    and demonstrate promising results on a QA task and the GLUE benchmark.
    
    \item Roof-Transformer addresses BERT's input length limitation.
	 We believe this will also facilitate NLP tasks where additional knowledge
	 or long context comprehension is needed.
    
	 \item We show that precise knowledge selection and effective representation
	 are critical to improve the performance for downstream NLP tasks.
\end{itemize}    

\section{Related Work}

Much work has been done to integrate knowledge bases or knowledge graphs for enhanced language
representation. 

Before strong pre-trained LMs such as BERT were proposed, research concerned
joint representation learning of words and knowledge. ~\cite{KG-2014}
combine knowledge embeddings and word vectors, and ~\cite{conv} utilize a
convolutional neural network to capture the compositional structure of textual
relations, and jointly optimize entity,
KBs, and textual relation representations. Both studies are based on the
concept of word2vec~\cite{word2vec} and TransE~\cite{TransE}. 

After Google Inc.\ launched BERT in 2018, studies on KB/KG integration
gradually focused on optimization with pre-trained LMs. ERNIE~\cite{ERNIE}, an
early study,  encodes knowledge information in KGs via TransE~\cite{TransE}, a knowledge
embedding model trained on Wikidata, and refines BERT
pre-training using named entity masking and phrase masking. K-BERT~\cite{KBERT-2019} 
injects knowledge into the text to form a sentence
tree 
without the need to pre-train a model    
for knowledge embeddings, and 
adopts soft-position embeddings and a visibility matrix for structural information
and to prevent diverting the sentence from its correct meaning. Based on
these works, KEPLER~\cite{KEPLER} jointly optimizes knowledge embeddings
and masks language modeling objectives on pre-trained LMs.

Other work involves the joint use of dual BERTs. For
example, Sentence-BERT~\cite{SBERT} uses a model to derive sentence
embeddings via BERTs, and uses a classifier to judge the similarity of two
sentences. DC-BERT~\cite{DC-BERT}, 
a decoupled contextual encoding framework to address the efficiency of
information retrieval, uses an online BERT to encode the question once, and
an offline BERT which pre-encodes each document and caches their encodings. 

There are also some solutions to sequence length limitation, such as BigBird~\cite{BigBird}, Longformer~\cite{Longformer}. Both of them address the quadratic dependency on sequence length by using sparse attention mechanism; while we adopt the full attention mechanism in our work.

\section{Methodology}
In this section, we describe in detail the framework of Roof-Transformer presented in
Fig.~\ref{fig:fig1}, including the input formats for Roof-Transformer, which are sentences
pairs and selected triples from KB.

\subsection{Model Architecture}
\label{sec:3-1}
As shown in Fig.~\ref{fig:fig1}, the model architecture of Roof-Transformer
contains three stacked modules: (1)~the underlying BERT model, responsible for
encoding tokens to meaningful representations; (2)~the Fusion Layer,
responsible for combining information from the underlying BERT model; and 
(3)~the Prediction Layer, responsible for downstream tasks---in our case,
QA and common natural language understanding (NLU) tasks.
\smallskip

\noindent \textbf{Underlying BERT model}\\
The underlying BERT model is composed
of two independent BERT models: TASK-BERT and KB-BERT. TASK-BERT
encodes the tokenized passages, which are identical to those input for a single
BERT on each downstream task, into embeddings. KB-BERT encodes the tokenized
triples from KBs to embeddings. Both embeddings are then concatenated and fed
to the fusion layer as input.
\smallskip

\noindent \textbf{Fusion layer}\\
We select Transformer Encoder (TE)~\cite{Transformer}, LSTM~\cite{LSTM} and Linear layer as the candidates of our Fusion layer. The
input of the fusion layer is the concatenation of the output embeddings from
TASK-BERT, $\mathit{Emb}\in \mathbb{R}^{M\times d}$, and the output embeddings
from KB-BERT, $\mathit{Emb}^{\prime} \in \mathbb{R}^{N\times d}$, where $d$ is the hidden dimension of the word embeddings, $M$ is the length of the tokenized question and paragraph, and $N$ is the length of the tokenized triples from the KBs.

\smallskip

\noindent \textbf{Prediction layer}\\
The Prediction Layer is simply a linear NN
layer, which is responsible for transforming high-dimension embeddings into
appropriate logits for prediction and inference. The input of the prediction
layer is the output embeddings of the fusion layer, $\mathit{Emb}^{\prime\prime}
\in \mathbb{R}^{(M+N)\times d}$, whereas the output of the QA task is
$\mathit{logits} \in \mathbb{R}^{(M+N)\times 2}$. The two dimensions of the
output logits in each position are the start and end logits, that is, 
the probability of whether the position is the start or the end position of the
answer. In other NLU tasks, the output embeddings of the fusion
layer are compressed to a single sequence length $\mathit{AvgEmb} \in
\mathbb{R}^{d}$, whereas the output is $\mathit{logits} \in \mathbb{R}^{e}$, where
$e$ is based on the number of classes of the predicted label.

The model parameters are updated by minimizing the cross-entropy loss between
the output logits and the ground truths.

\subsection{TASK-BERT Input Format}
In this paper, we test the capability of Roof-Transformer on a QA downstream task and
the GLUE NLU tasks. We follow the format of the input of the major approach for
BERT. Each sentence pair from the dataset contains two passages, both of which
are tokenized and concatenated with a \sep token for input to TASK-BERT.

Since BERT has a 512-token limitation on input length, we set a maximum length for
our question passage and paragraph passage. If the length of the passage is
shorter than the maximum length, which is generally the case in the question
passage, \pad tokens are appended to the concatenated passage to
fix the length of every input. If the passage length exceeds the
maximum length, which is generally the case in paragraph passages, we 
truncate the passage.

\subsection{KB-BERT Input Format}
\label{sec:3-3}
For our approach, we choose KBs as our external information. The KB contains triples,
each of which consists of a head, a relation, and a tail, 
corresponding to the subject, the relation, and the object in a sentence.

The triples are selected via a heuristic algorithm (string match), which
selects a triple if its head exists in the paragraph passage in the TASK-BERT input.
The selected triples are then concatenated together and separated by the \sep token.

\begin{figure}[t]
    \hspace{-0.5em}\includegraphics[width=1.05\columnwidth]{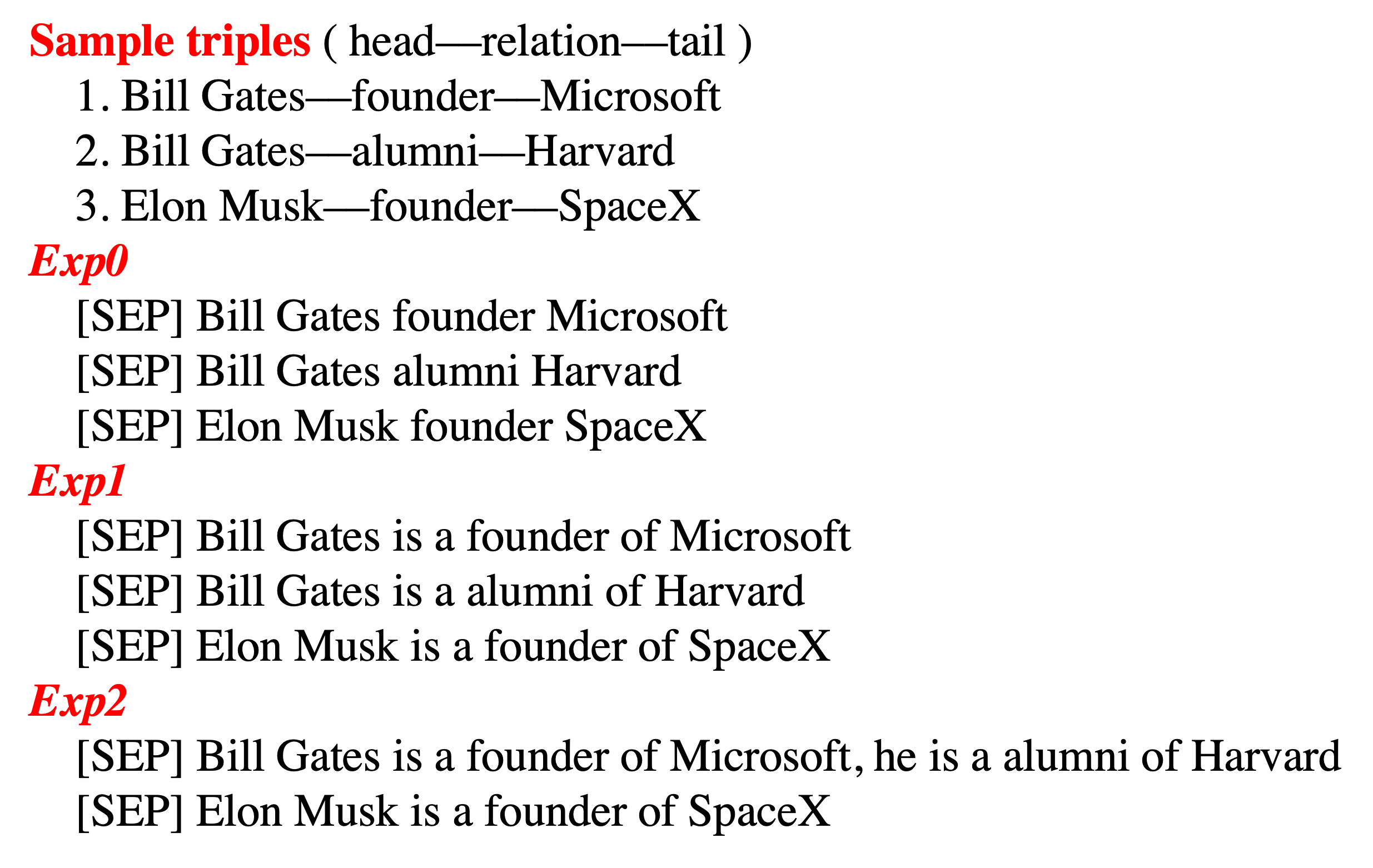}
    \caption{Three expansion types.}
    \label{fig:fig2}
\end{figure}

As shown in Fig.~\ref{fig:fig2}, we propose three expansion 
types---\textit{Exp0}, \textit{Exp1}, and \textit{Exp2}---for the selected triples.
\textit{Exp0} simply concatenates the components in the triple as a unit,
  and appends this to    the previous unit.   
For Chinese KBs, \textit{Exp1} further
adds \textsc{[的]}, a Chinese token, between the head and relation, and adds \textsc{[是]} between the relation and tail to form a natural sentence~\cite{KG-Syn}. For English KBs, the input format will be equivalent to ``head is a relation of tail''. The sentence, which is also the unit, is then concatenated after the previous unit.
\textit{Exp2} is a refined version of \textit{Exp1}: if the current
head of the selected triple is identical to the head of the previous unit, the
head is replaced by a pronoun and merged with the previous unit with a
comma to form a larger sentence/unit.

\section{Experiments}
\label{sec4}
In this section, we describe 
  the Roof-Transformer training           
as well as the
fine-tuning results with different KBs and settings. The estimated number of
parameters ranges from 200M to~230M, depending on the depth and of candidate the fusion layer. (see Appendix~\ref{sec:7-2-parameter} for parameter setting details)

\subsection{Dataset}
For the QA task, we evaluated Roof-Transformer and the baseline models (Sec.~\ref{sec:4-6-baseline}) on the DRCD 
dataset~\cite{DRCD}, a Chinese QA benchmark. For the NLU tasks, we evaluated the model on the General Language
Understanding Evaluation (GLUE) benchmark~\cite{wang2018glue}, which was used for
ERNIE~\cite{ERNIE}. GLUE is an English multi-task NLU benchmark consisting of 11~tasks,
of which we used 8 to evaluate Roof-Transformer and compare it
with the baseline models. These tasks use different evaluation metrics depending
on their purpose. (see Appendix~\ref{sec:7-1-data} for statistical details of each dataset)
\smallskip

\noindent \textbf{DRCD}\\
The Delta Reading Comprehension Dataset is an
open-source Chinese QA dataset composed of paragraphs from Wikipedia
articles and questions generated by annotators. The ground truths of each
question-paragraph pair are the start and end position of the answer. The
result is evaluated by the exact match (EM) score.
\smallskip

\noindent \textbf{GLUE}\\
We selected the following eight English GLUE benchmark tasks: 
(1)~SST-2, a sentiment task using accuracy as the metric; (2)~CoLA, an acceptability
task using Matthews correlation; (3)~MRPC, a paraphrase task using the F1
score; (4)~STS-B, a sentence similarity task using Pearson--Spearman
correlation; (5)~QNLI, a natural language inference (NLI) task using
accuracy; (6)~QQP, a paraphrase task using the F1 score;
(7)~RTE, a NLI task using accuracy; and (8)~MNLI, a NLI task
using accuracy.

\subsection{Knowledge Base}
We employ HowNet and CN-DBpedia, Chinese KBs which are refined and used in K-BERT~\cite{KBERT-2019} for QA. Each triple in the KBs includes a head,
a relation, and a tail. We also use the KELM corpus as English KB to evaluate the model on common
NLU tasks in the GLUE benchmark.
\smallskip

\noindent \textbf{CN-DBpedia}\\
CN-DBpedia~\cite{Cn-Dbpeida} is a
large-scale structured encyclopedia developed and maintained by the Knowledge
Workshop Laboratory of Fudan University. It has been extended to fields such as law,
industry, finance, and medical care, providing supporting knowledge services
for intelligent applications in various industries. We use a refined version of CN-DBpedia by eliminating triples whose entity names are less than 2 in length or contain special characters like what was done in K-BERT. The refined CN-DBpedia
contains around 5M triples. 
\smallskip

\noindent \textbf{HowNet}\\ 
HowNet~\cite{HowNet} is a large-scale KB
containing Chinese concepts and vocabulary. Each entity is annotated with
semantic units called sememes. In Hownet, sememes refer to some basic unit of senses. The triples in HowNet can be represented as {head, contain, sememes}. We adopt the same method in CN-DBpedia to obtain the refined version of HowNet, which contains a total of 52,576 triples.
\smallskip

\noindent \textbf{KELM}\\
The KELM corpus~\cite{lu2021kelm} consists of the
entire Wikidata KG as natural text sentences. It contains around 15M sentences converted from KG's triples. The sentences are like the \textit{Exp2} examples in Fig.~\ref{fig:fig2}.


\subsection{KB Format}
\label{sec4.3}
Knowledge representation clearly effects model performance 
by influencing the differentiation quality of language understanding. 
The knowledge selected also effects model performance. Therefore, we evaluated 6
different input formats for KB-BERT, that is, combinations of 3 kinds of
knowledge representation with 2 kinds of knowledge selection.
\smallskip

\noindent \textbf{Representation}\\
We represented KB knowledge with 3
types of expansions as mentioned in Section~\ref{sec:3-3} (\textit{Exp0, Exp1, Exp2}) and demonstrated in
Fig.~\ref{fig:fig2}, and evaluated the model performance using these representations.
\smallskip

\noindent \textbf{Selection}\\
As mentioned in Section~\ref{sec:3-3}, a KB triple
is selected if its head exists in the paragraph passage, regardless of whether its tail
exists in the paragraph: this is denoted as as \textit{No\_Tail}. The
other kind of selection is \textit{Has\_Tail}, meaning that the selected triple's head
and tail are both present in the paragraph passage. If the tail exists in 
the paragraph passage, the selected triple is more likely to be related to the paragraph passage.


\subsection{KB Encoder}
Since the computational complexity of self-attention mechanism is \emph{O(N$^{2}$)}, where \emph{N} is the length of input tokens, acquiring knowledge embeddings can be computationally expensive. Hence, apart from using KB-BERT with full-attention, we also investigate the performance of our model with \emph{cached} KB-BERT.

In cached KB-BERT experiment, we treated KB-BERT as an off-line BERT by acquiring knowledge embeddings through a cached (freeze) BERT before training, and the embeddings are then concatenated with the output of Task-BERT forming the input of the fusion layer during training and inference. It is worth noting that different from inputting the concatenated knowledge to KB-BERT all at once, we feed each triple to cached KB-BERT to get its embedding and dynamically concatenate all triples' embeddings to feed to the fusion layer. For example, take \texttt{\cls Bill Gates is a founder of Microsoft \sep Elon Musk is a founder of SpaceX \sep} as the original input; thus the individual input of cached KB-BERT will be \texttt{\cls Bill Gates is a founder of Microsoft \sep} and \texttt{Elon Musk is a founder of SpaceX \sep}; then both output embeddings will be concatenated to form the knowledge embeddings as part of the input of the fusion layer (the other part is the passage embeddings from TASK-BERT).


\subsection{Other Setting}
\label{sec4.5}
We also conducted experiments to investigate the following factors which influence fusion efficiency.

\noindent \textbf{Segmentation}\\
Since the input of the fusion layer contains outputs from both underlying BERT
models, each of which contains its own positional information, it is difficult 
for the Fusion Layer to distinguish the two parts. Thus, we add segmentation tokens to the
KB-BERT and TASK-BERT input, and evaluate the two formats, which are denoted as \textit{type-1} and \textit{type-2} segmentation, to determine
which is more effective.

The \textit{type-1} segmentation of KB tokens and padding tokens in
KB-BERT is different, using \texttt{[A]} and \texttt{[B]} respectively; in TASK-BERT, the segmentation of
question, paragraph, and padding tokens is \texttt{[A]}, \texttt{[B]}, \texttt{[A]} respectively.
The first type of segmentation aims to separate the content of tokens in a
single BERT.

As shown in Fig.~\ref{fig:fig1}, in the \textit{type-2} segmentation, every token in KB-BERT is set to \texttt{[A]}; whereas in TASK-BERT, the segmentation of every token is
set to \texttt{[B]}. With the
second type of segmentation we seek to separate the content of tokens of the two BERTs.




\smallskip

\noindent \textbf{Fusion layer}\\
We test different (1) Candidates (Linear, LSTM and Transformer encoder), (2) Depth, (3) Initialization (pre-trained or not) and (4) Learning rate of Fusion Layer.

It is worth noting that we modify the learning rate of Fusion Layer by increasing it comparing to those of Underlying BERTs, e.g. 5, 10, 20 times of the Underlying BERTs' learning rate due to different model complexities of BERTs (high) and Fusion Layer (low).

\smallskip
\noindent The investigations in Sec.~\ref{sec4.3} - ~\ref{sec4.5} are not only for better performance but also aim to answer the the questions mentioned in Sec.~\ref{sec1}.


\begin{table*}[t!]
    \begin{center}
    \scalebox{0.85}{
        \begin{tabular}{lccccc}
            \toprule
            \bf Model & \bf \thead{Max \\ Para. Len.} & \bf \thead{Max \\ Know. Len.} & \bf KB  & \bf \thead{Fusion Layer \\ Init.}  & \bf EM score \\
            \midrule
            BERT$_{base}$-chinese & 450 & 0  & - & - & 76.08 \\
            \midrule \midrule
            BERT$_{base}$-chinese$^{\dagger}$ & 420 & 30 & HowNet & - & 76.20 $\uparrow$ \\
            BERT$_{base}$-chinese$^{\dagger}$ & 400 & 50 & HowNet & - & 75.94 $\downarrow$ \\
            \midrule
            BERT$_{base}$-chinese$^{\dagger}$ & 420 & 30 & CN-DBpedia & - & 75.63 $\downarrow$ \\
            BERT$_{base}$-chinese$^{\dagger}$ & 400 & 50 & CN-DBpedia & - & 76.07 $\downarrow$ \\
            \midrule \midrule
            Roof-Transformer  & 450 & 511 & HowNet & BERT$_{base}$-chinese &  77.45 $\uparrow$ \\
            Roof-Transformer  & 450 & 511 & CN-DBpedia & BERT$_{base}$-chinese & \textbf{77.59} $\uparrow$ \\
            \midrule
            Roof-Transformer  & 450 & 511 & HowNet & random weight & 76.31 $\uparrow$ \\
            Roof-Transformer  & 450 & 511 & CN-DBpedia & random weight & 76.61 $\uparrow$ \\
            \bottomrule
        \end{tabular}}
    \caption{\label{tab:QA}Results of Roof-Transformer and baselines on QA tasks (\%) with different KBs, initialization (init.), max paragraph length (para. len.) and max knowledge length (know. len.). Note that $^{\dagger}$ indicates using knowledge in single BERT architecture (baseline).}
    \end{center}
\end{table*}

\begin{table*}[t!]
    \begin{center}
    \scalebox{0.85}{
    \begin{tabular}{lccccccccc}
        \toprule
        \bf Model & \bf KB &\bf SST-2 & \bf CoLA & \bf MRPC & \bf STS-B & \bf QNLI & \bf QQP & \bf RTE & \bf MNLI-m \\
        \midrule
         BERT$_{base}$ & - & 93.3 & 52.1 & 88.0 & \bf{85.0} & 90.5 & \bf 71.2 & 66.4 & \bf 84.6 \\
         ERNIE  & Wikidata &\bf 93.5 & 52.3 & 88.2 & 83.2 & \bf 91.3 & 71.2 & 68.8 & 84.0\\
        \midrule 
         Roof-Transformer & KELM & 93.0 & \bf 54.4 & \bf 89.0 & 84.2 & 90.6 & 70.3 & \bf 69.0 & 84.3 \\
        \bottomrule
    \end{tabular}}
    \caption{\label{tab:glue}Results of Roof-Transformer and baselines on eight datasets of GLUE benchmark (\%)}
    \end{center}
\end{table*}

\vspace{-1.0mm}
\subsection{Baseline}
\label{sec:4-6-baseline}
In QA task, we compare Roof-Transformer to two baselines: BERT$_{base}$-chinese \cite{devlin2018bert} without knowledge and a single BERT$_{base}$-chinese with knowledge. In NLU tasks of GLUE, we compare Roof-Transformer to two baselines: BERT$_{base}$ without knowledge,  and ERINE \cite{ERNIE}.

BERT$_{base}$-chinese is pre-trained on WikiZh; BERT$_{base}$, is pre-trained on the BookCorpus and English Wikipedia; ERNIE is pre-trained on English Wikipedia for large-scale textual corpora and Wikidata for KGs.


\begin{figure*}[t!]
    \makebox[\textwidth][c]{\hspace{-0em}\includegraphics[width=0.87\textwidth]{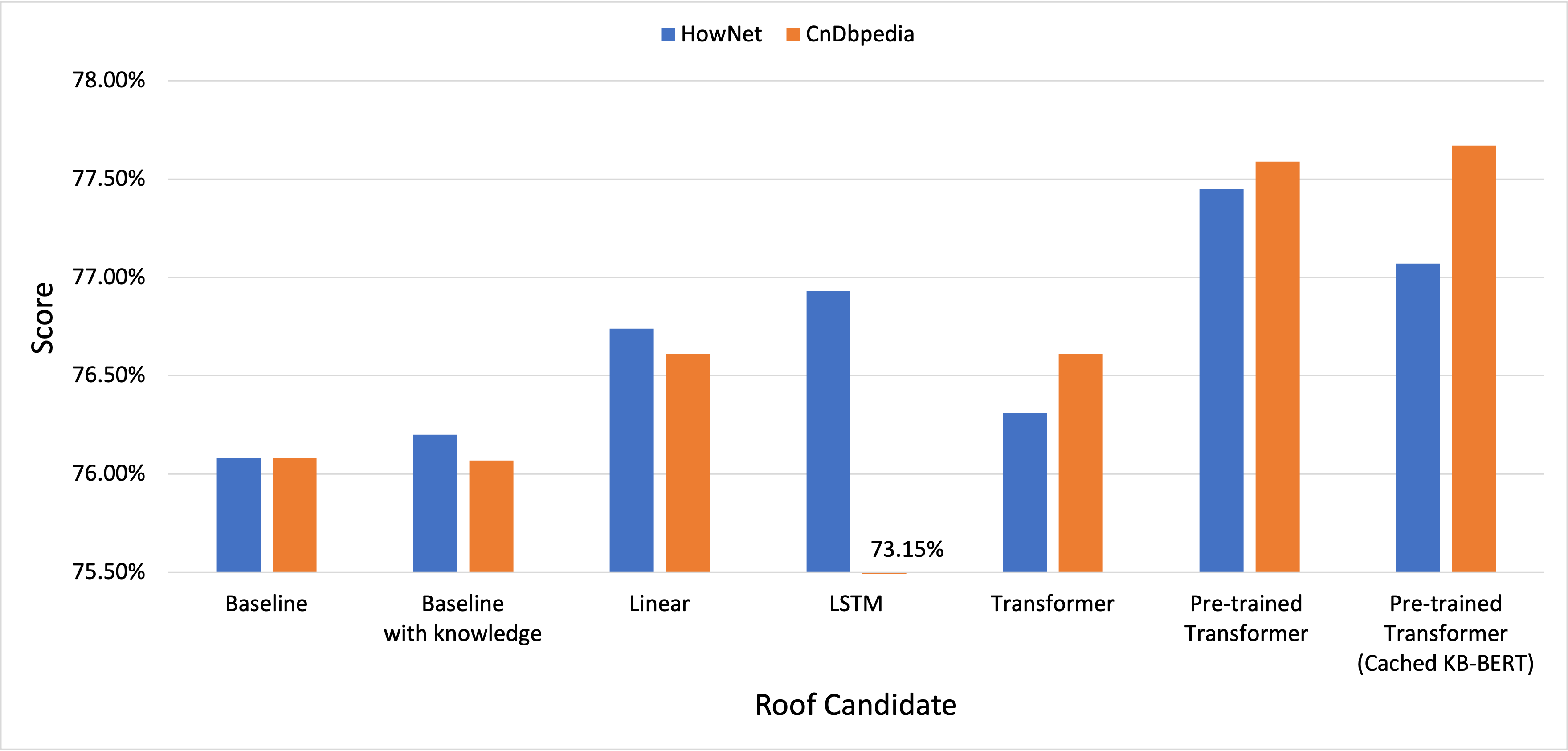}}
  \vspace{-7mm}
  \caption{Comparison of performance of different roofs in QA task.}
  \label{fig:fig3}
\end{figure*}

\begin{figure*}[t!]
    \makebox[\textwidth][c]{\hspace{-0em}\includegraphics[width=0.87\textwidth]{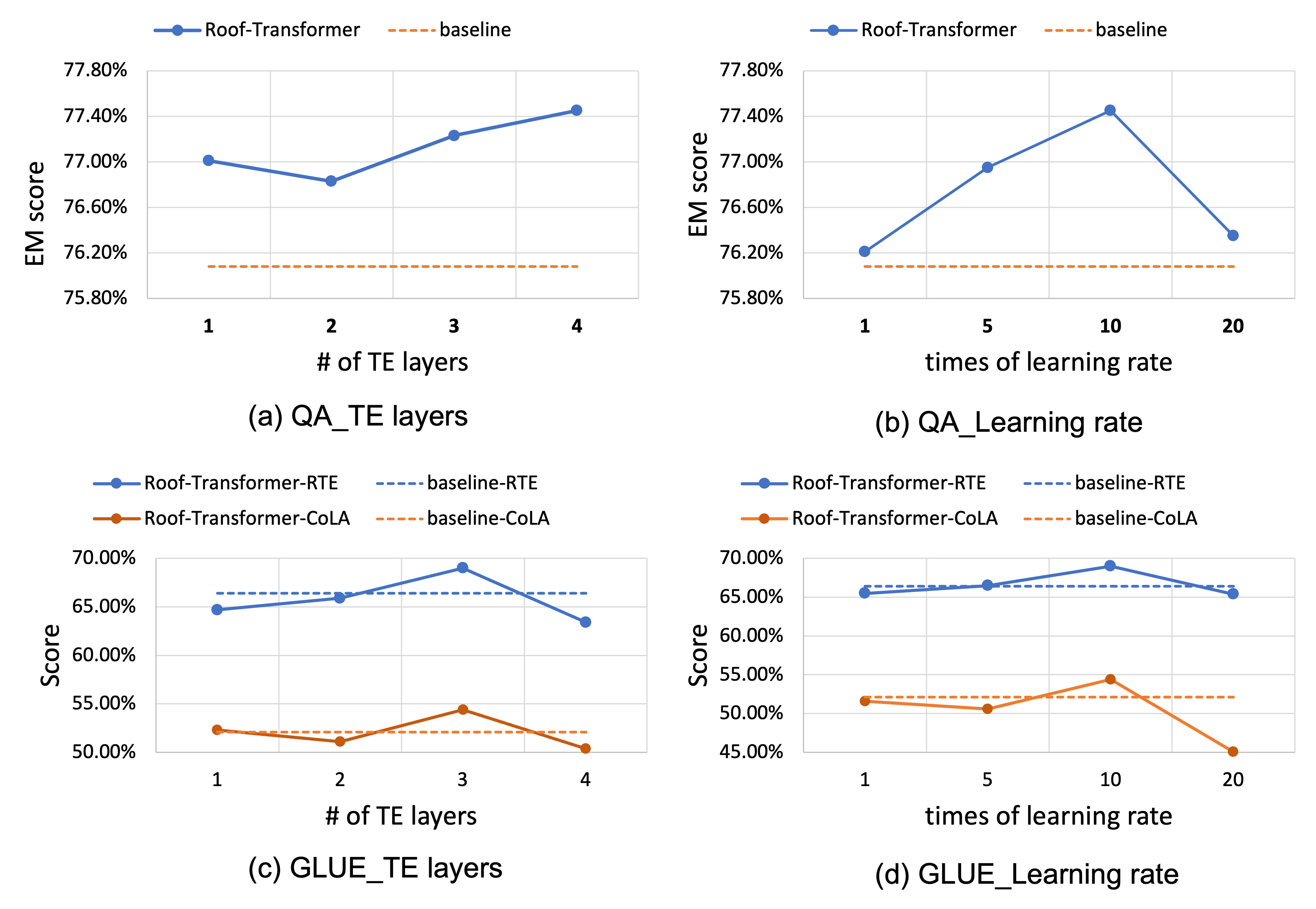}}
  \vspace{-7mm}
  \caption{Fusion efficiency results on QA and GLUE tasks: The metrics
  in (a) and (b) are both EM scores, whereas the metrics in (c) and (d) are accuracy
  for RTE and Matthew's correlation for CoLA. In (a) and (c), we use 10 times the learning
  rate of the underlying BERT model in the TE layers; in (b), we set the number of
  TE layers to 4; in (d), we set the number of TE layers to 3.}
  \label{fig:fig4}
\end{figure*}

\subsection{Results}
\label{sec:4-7-QA}

\noindent \textbf{Necessity of Dual BERT}\\
As mentioned in Sec.~\ref{sec:4-6-baseline}, for QA task, we test baseline with knowledge which encodes both text and knowledge with single BERT and without Fusion Layer. That is, a partition of paragraph should be sacrificed for knowledge tokens. (see Appendix~\ref{sec:7-2-parameter} for details of input length settings)

As we can see in Table~\ref{tab:QA}, sacrificing information of paragraph in exchange of knowledge may be slightly beneficial in some cases; however, in most cases, it results in performance drop. Therefore, the need of second BERT to encode knowledge separately is guaranteed.
\smallskip


\noindent \textbf{Candidate of Fusion Layer}\\
As shown in Figure~\ref{fig:fig3} (all settings are the same and follow Appendix~\ref{sec:7-3-exp}), the performance of every roof candidates beats both baselines (except a dramatic drop in LSTM with the use of CN-DBpedia). Linear and LSTM both achieve better performance compared to Transformer without pre-train while using HowNet. The pre-trained Transformer consistently outperforms its variants with different KBs (the Cached KB-BERT case will be discussed later). This shows the effectiveness of our roof-architecture and the power of pre-trained transformer, which was selected as our Fusion Layer for the rest of the experiments.
\smallskip

\noindent \textbf{QA performance}\\
The best parameter setting for QA task we found through experiments are presented in Appendix~\ref{sec:7-3-exp}.
As shown in Table~\ref{tab:QA}, with external information
from KBs, the EM score improvement on the QA task exceeds
1.5\%, which attests the benefits of utilizing KBs. 
Perhaps the slight difference between the EM scores of CN-DBpedia and HowNet in Table~\ref{tab:QA} is because CN-DBpedia has better and more knowledge than HowNet.
\smallskip

\noindent \textbf{KB format}\\
In Table~\ref{tab:format} we compare the results of the 6 different KB formats (setting follows Appendix~\ref{sec:7-3-exp} except KB format). The KB format with \textit{Exp2}~+~\textit{Has\_Tail} clearly yields
the best performance. However, a closer look at the contributions of
knowledge representation and selection shows that \textit{Exp2} yields
marginally better results among three expansion types, producing finer language representation; even more crucial, however, is whether the tails of the selected triples are in the paragraph passage.
The knowledge selection comparison is shown in Table~\ref{tab:tail} (setting mentioned in Appendix~\ref{sec:7-3-exp} except knowledge selection). The results indicate that selecting \textit{Has\_Tail} knowledge---with tails in the paragraph passage---improves the EM score even with far less knowledge, suggesting that adding arbitrary information could harm performance.

\smallskip

\noindent \textbf{Depth of Fusion Layer}\\
As reported in Figures~\ref{fig:fig4}(a)
and~\ref{fig:fig4}(c), the scores peak when using the last 4 and 3 TE layers from the
pre-trained BERT as the Fusion layer. Using more layers consumes excessive
memory and yields lower scores, perhaps due to overfitting; when using
too few layers, the fusion layer is unable to effectively integrate language 
representation with the KBs.
\smallskip

\begin{table}[t]
    \begin{center}
    \scalebox{0.9}{
    \begin{tabular}{l|cc}
        \toprule
        \bf Type & \bf \textit{Has\_Tail}  & \bf \textit{No\_Tail}    \\
        \midrule
        \bf \textit{Exp0} & 77.21 & 77.34  \\
        \bf \textit{Exp1} & 76.92 & 76.77  \\
        \bf \textit{Exp2} & \bf 77.45 & 77.03  \\
        \bottomrule
    \end{tabular}}
	 \caption{\label{tab:format} EM scores (\%) for 6 types of KB
	 selection and representation on QA task using HowNet as KB.}
    \end{center}
\end{table}

\begin{table}[t]
    \begin{center}
    \scalebox{0.85}{
    \begin{tabular}{lccc}
        \toprule
        \bf KB & \bf Selection  & \bf Length & \bf EM score (\%)\\
        \midrule
        HowNet & \textit{Has\_Tail} & 29 & \bf 77.45 \\
        HowNet  & \textit{No\_Tail} & 131 &  77.03 \\
        \midrule
        CN-DBpedia  & \textit{Has\_Tail} & 24 & \bf 77.59 \\
        CN-DBpedia & \textit{No\_Tail} & 168 & 76.90 \\
        \bottomrule
    \end{tabular}}
	 \caption{\label{tab:tail} Results with different knowledge
	 selections, where Length is the average length of the input tokens of
	 KB-BERT (w/o \pad) during training.}
    \end{center}
\end{table}

\noindent \textbf{Learning rate of Fusion Layer}\\
As reported in Fig.~\ref{fig:fig4}(b), using 10 times the learning rate of the underlying BERT model’s learning rate yields the highest EM
score, for an increase of over 1\% compared to using the same learning rate as the underlying BERT model. Similarly, as reported in Fig.~\ref{fig:fig4}(d), using 10 times learning rate yields an increase of 2\%. This shows that a higher learning rate in the fusion layer improves the fusion effectiveness and further enhances prediction.
This is because after BERT is pre-trained, a small
learning rate is sufficient for fine-tuning on downstream tasks, and requires fewer epochs for its loss to converge to a minima (global or local); by contrast, a higher learning rate could prevent the model from converging. 
Thus it may be that the learning pace of the fusion layer should be
faster than that of the underlying BERT model in order to 
make best use of the information provided by each underlying BERT instead of being confined to one or the other.
\smallskip

\noindent \textbf{Cached KB-BERT}\\
As shown in Figure~\ref{fig:fig3}, cahcing KB-BERT does not lead to performance drop but even slightly improve around 0.1\% (77.67\%) with the use of Cn-Dbpedia comparing to our original best setting. A possible explanation is that without full self-attention between knowledge triples, it could reduce noise during encoding the triples. Moreover, without updating parameters of KB-BERT during fine-tuning, it reduces 33\% of the computational cost, making prediction and inference more efficient.
\smallskip

\noindent \textbf{GLUE}\\
The best parameter setting for NLU tasks are presented in Appendix~\ref{sec:7-3-exp}. As shown in Table~\ref{tab:glue}, Roof-Transformer outperforms
datasets like RTE, CoLA, and MRPC, which shows that the
proposed model effectively integrates knowledge and contextual embeddings.
Moreover, Roof-Transformer achieves comparable or even superior results with ERNIE on the
GLUE benchmark.

The results also show that Roof-Transformer has no significant effect on 
tasks like SST-2, STS-B, and QNLI. It is probably due to the need of
external information for tasks like sentimental analysis: sentence sentiment
is determined by emotional words but not knowledge. This phenomenon is
reflected also in studies like ERNIE and K-BERT. However, for these tasks, the proposed
model still achieves performance comparable to that of BERT. For tasks that our model beats both ERINE and BERT like RTE, we have conducted case study to further investigate how knowledge effects prediction.

\subsection{Case Study}
We conduct case studies on DRCD dataset and RTE dataset in GLUE benchmark to find out how knowledge helps or misleads our model. We select three examples from dev (validation) data where two of them are positive samples (our model answers correctly, while the baseline answers incorrectly) and one of them is a negative samples (our model answers incorrectly, while the baseline answers correctly) in both case studies. The details of examples and analysis are presented in Appendix~\ref{sec:7-4-case}.

\section{\centering{Conclusion}}
In this paper, we propose Roof-Transformer to encode knowledge and input sentences using two underlying BERTs with a fusion layer (transformer encoder) on top. This architecture relaxes BERT's length limitation, allowing BERT to use more knowledge and longer input texts. Experimental results on a QA task and the GLUE benchmark demonstrate the model's effectiveness. We also show that precise knowledge selection is critical under this architecture. Roof-Transformer is a general and powerful method for language understanding which can easily be applied to other NLP tasks. It is likely to especially benefit tasks which require information from a large range of knowledge or long input texts. For future work, we plan to replace the underlying BERTs to other pre-trained LMs, such as RoBERTa~\cite{RoBERTa} and XLNet~\cite{XLNET} to to consolidate the effectiveness of the proposed framework.

\bibliography{anthology,custom}
\bibliographystyle{acl_natbib}

\newpage
\section{Appendices}
\label{sec:appendix}

\subsection{Dataset Statistics}
\label{sec:7-1-data}

\begin{table}[h]
    \begin{center}
    \scalebox{0.85}{
    \begin{tabular}{lccc}
        \toprule
        \bf Dataset & \bf Train  & \bf Validation & \bf Test \\
        \midrule
        \multicolumn{4}{c}{\textbf{QA task}}\\
        DRCD & 26,935 & 3,523 & 3,492 \\
        \midrule
        \multicolumn{4}{c}{\textbf{GLUE benchmark}}\\
        SST-2 & 67,349 & 872 &  1,821 \\
        CoLA  & 8,551 & 1,043 & 1,063 \\
        MRPC & 3,668 & 408 & 1,725 \\
        STS-B & 5,749 & 1,500 & 1,379 \\
        QNLI & 104,743 & 5,463 & 5,463 \\
        QQP & 363,846 & 40,430 & 390,965 \\
        RTE & 2,490 & 277 & 3,000 \\
        MNLI-m & 392,702 & 9,832 & 9,847 \\
        \bottomrule
    \end{tabular}}
    \caption{\label{tab:table5} Number of instances in train-dev-test split of different datasets.}
    \end{center}
\end{table}

\subsection{Parameter Settings}
\label{sec:7-2-parameter}
For QA, given BERT's input length limit, we set the maximum
length of the question and paragraph to 59 and 450, respectively so that the total length of the input token in Task-BERT would be len(\cls) + len(Question tokens) + len(\sep) + len(Paragraph tokens) + len(\sep) = 1 + 59 + 1 + 450 + 1 = 512. Similarly, we set the maximum length of the knowledge as 511, so that the total
length of the input token in KB-BERT would be len(\cls) + len(Knowledge tokens) = 1 + 511 = 512.

The following setting values were found suitable for the QA datasets: a batch
size of 16 and an AdamW learning rate of $3e^{-5}$. In addition, we used the linear
learning rate decay scheduler. For the epoch count, as the
fine-tuning was for a downstream QA task, the epoch count was set to 1
with the training loss and accuracy converging properly.

For the GLUE NLU tasks, we set the maximum length of the sentence pairs to 512,
including one \cls token and two \sep tokens. If the sentence length 
was less than 512, we appended \pad tokens to the sentences until
the sentence length was 512.

The following setting values were found suitable for the GLUE benchmark:
a batch size of 16 and a learning rate of $2e^{-5}$. We used the cosine learning rate
decay scheduler. The training epochs were set to 5 for fine-tuning.

\subsection{Experiment Settings}
\label{sec:7-3-exp}
For QA task, we find the following setting performs the best in Roof-Transformer: (1) KB format with \textit{Exp2} and \textit{Has\_Tail}, (2) \textit{type-2} segmentation, (3) using pre-trained Transformer encoder layer (last k layers from BERT$_{base}$-chinese) for Fusion Layer, (4) 4 Fusion Layers (k = 4) , (5) using 10 times of the Underlying BERTs model's learning rate in Fusion Layer. 

For NLU tasks of GLUE, only the number of Fusion Layers is found to be 3 (k = 3) for best performance; other settings remain the same as QA task.

\subsection{Case Study}
\label{sec:7-4-case}
\noindent \textbf{QA task}\\
The Question-Answering is the task of answering a Question from a given Paragraph.

\begin{figure*}[t!]
      \makebox[\textwidth][c]{\hspace{-0em}\includegraphics[width=1\textwidth]{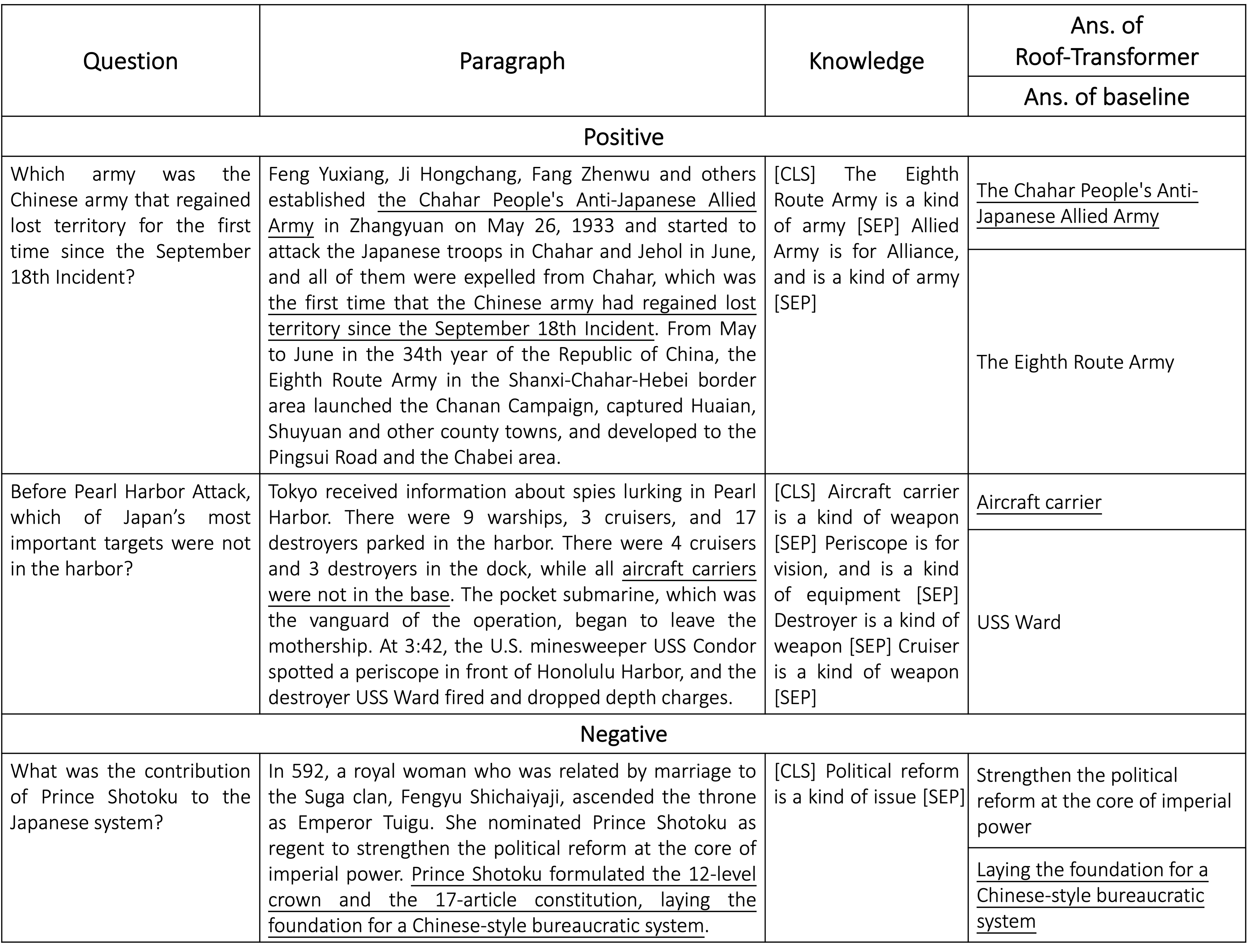}}
    \caption{Case study of QA task: Answers with underlines are also the ground truth of those examples. The answers of Roof-Transformer are align with the ground truth in positive examples, whereas answer of baseline is align with the ground truth in negative examples. Note that the content is translated (without loss in meaning), since the DRCD dataset and HowNet are in Chinese, and only a part of paragraph and knowledge are present for clarity.}
      \label{fig:fig5}
    \end{figure*}

As shown in Fig.~\ref{fig:fig5}. In the first positive example, knowledge provides information that both the Chahar People's Anti-Japanese Allied Army and the Eighth Route Army are kinds of armies, enabling Roof-Transformer to understand these unseen named entities, which indirectly aids prediction. Similarly, in the second positive example, Roof-Transformer also benefits from the added knowledge, which makes it aware of the functions and
character of the various named entities (aircraft carrier, destroyer, \ldots) in the paragraph. In contrast, without KB knowledge, the baseline BERT$_{\mathit{base}}$-chinese fails to understand the contextual information of the given question and paragraph, leading to misprediction.

On the other hand, in the negative example, the selected knowledge indicates that political reform is a kind of issue, which subtly response to the word "contribution" asked in the question, resulting in a biased prediction. On the contrary, without the information of knowledge, the baseline model is able to answer correctly.
\smallskip
\smallskip

\noindent \textbf{RTE task}\\
The Recognizing Textual Entailment is the task of determining whether the meaning of the Hypothesis is entailed (can be inferred) from given Text.

\begin{figure*}[t!]
      \makebox[\textwidth][c]{\hspace{-0em}\includegraphics[width=1\textwidth]{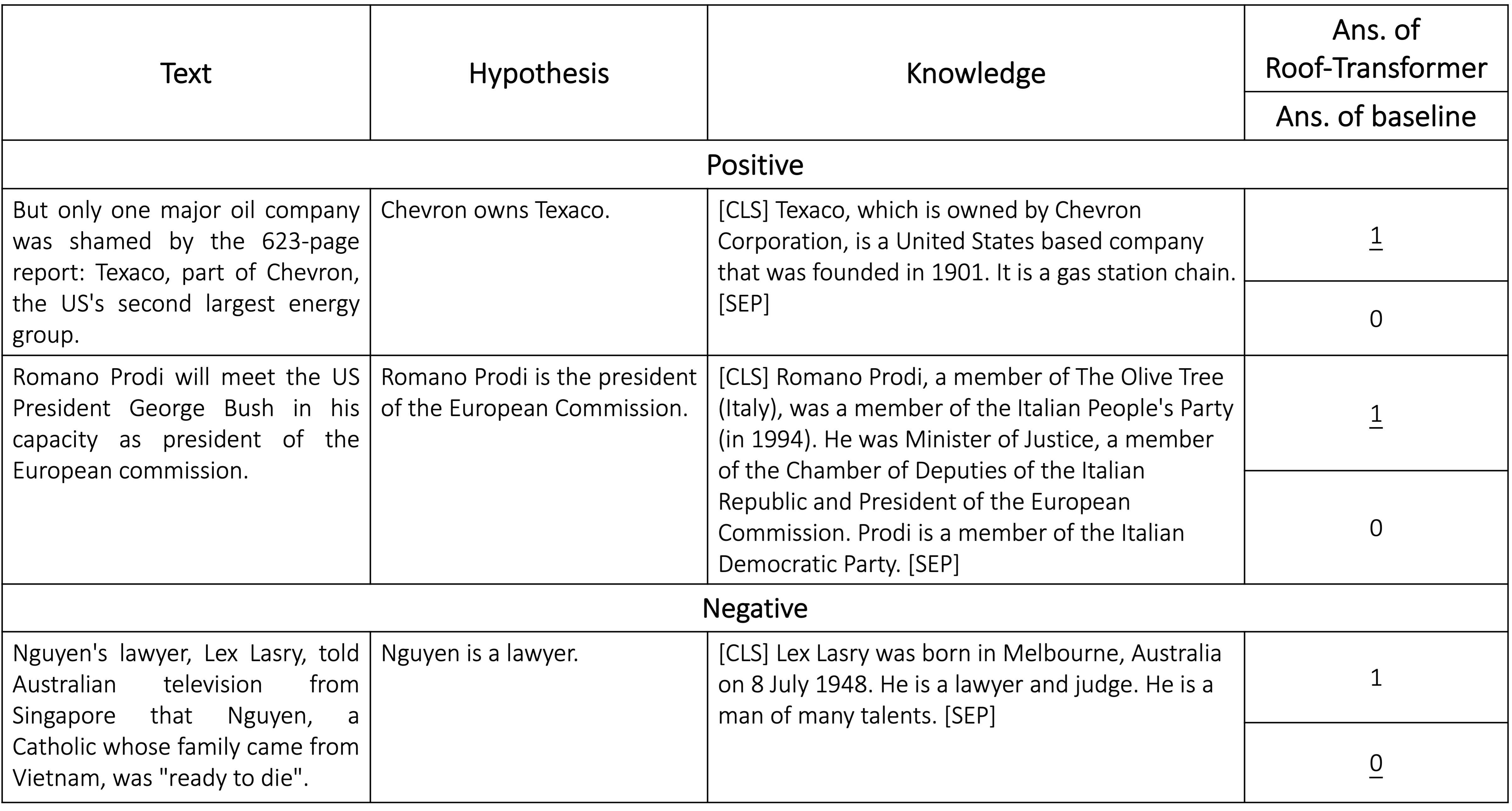}}
      \caption{Case study of RTE task: Similarly, the answers with underlines are the ground truth of those examples. Note that the content is originally in English, and only a part and knowledge is present for clarity.}
      \label{fig:fig6}
    \end{figure*}

As shown in Fig.~\ref{fig:fig6}. In the first positive example, the knowledge tells the fact of "Texaco is owned by Chevron Corporation", which helps our model to infer that sentence 1 entails sentence 2 instead of telling model the answer. Similarly, in the second positive example, Roof-Transformer also benefits from the fact related to Romano Prodi that knowledge provides.  In contrast, without the help of knowledge corpus, the baseline BERT$_{\mathit{base}}$ fails to make inference only based on sentence 1 and sentence 2.

On the other hand, in the negative example, although the selected knowledge tells fact about "Lex Lasry", there exista a huge overlap between knowledge and sentence 2 (i.e., "is a lawyer"), resulting in a misprediction. On the contrary, without the information of knowledge, the baseline model is able to make correct inference.

\smallskip

\end{CJK}
\end{document}